\documentclass[letterpaper, 10 pt, conference]{ieeeconf}  % Comment this line out if you need a4paper

\IEEEoverridecommandlockouts                              % This command is only needed if 
\usepackage{cite}
\usepackage[utf8]{inputenc} % allow utf-8 input
\usepackage[T1]{fontenc}    % use 8-bit T1 fonts
\usepackage{hyperref}       % hyperlinks
\usepackage{url}            % simple URL typesetting
\usepackage{booktabs}       % professional-quality tables
\usepackage{amsfonts}       % blackboard math symbols
\usepackage{nicefrac}       % compact symbols for 1/2, etc.
\usepackage{microtype}      % microtypography
\usepackage{xcolor}         % colors
\let\labelindent\relax
\usepackage{enumitem}

\usepackage{array}
\newcolumntype{C}[1]{>{\centering\arraybackslash}p{#1}}
\usepackage{makecell}

\usepackage{graphicx}
\usepackage{wrapfig}
\usepackage{pgfplots}
\usepackage{float}
\usepackage{tcolorbox}
\usepackage{multirow}
\usepackage{subcaption}
\usepackage{CJK}
\usepackage{algorithm}
\usepackage{algorithmicx}
\usepackage{algpseudocode}
\usepackage{amsmath}
\usepackage{adjustbox}
\usepackage{pifont}
\usepackage{amssymb}
\usepackage{makecell}

\usepackage[table]{xcolor}

\usepackage{longtable}
\usepackage{geometry}
\floatname{algorithm}{algorithm}

\newcommand{\xmark}{\textcolor{red!40!black}{\ding{55}}}
\newcommand{\cmark}{\textcolor{green!60!black}{\ding{51}}}
\newcommand{\projname}{DualTHOR}
\newcommand{\projnamebold}{\textbf{\projname}}
\newcommand{\aithor}{AI2-THOR}

\definecolor{lightgray}{gray}{0.9}
\definecolor{mydarkblue}{rgb}{0,0.08,0.45}
\definecolor{mydarkgreen}{RGB}{0, 139, 69}
\definecolor{mygreen2}{RGB}{0 205 0}
\definecolor{mybrown}{RGB}{139 69 19}
\definecolor{mypink}{RGB}{184 131 211}
\definecolor{myblue2}{RGB}{187 151 39}
\definecolor{boxblue}{RGB}{79,173,234}
\definecolor{boxgreen}{RGB}{159,206,99}
\definecolor{tablepeach}{RGB}{251, 240, 217}
\definecolor{tablepurple}{RGB}{248,235,252}
\definecolor{tableblue}{RGB}{235,241,255}
\definecolor{lowerbody}{RGB}{76,123,49}
\definecolor{upperbody}{RGB}{47,110,186}

\definecolor{lightblue}{HTML}{7FB2D3}
\definecolor{lightgreen}{HTML}{90B56D}  
\definecolor{darkblue}{HTML}{367DB0}  
\definecolor{darkgreen}{HTML}{3D9F3C} 
\hypersetup{
    colorlinks=true,
    breaklinks=true,
    citecolor={green!70!black},
}

\title{Towards Proprioception-Aware Embodied Planning for Dual-Arm Humanoid Robots}

\author{%
  \textbf{Boyu Li$^{1,2,3}$, Siyuan He$^{4}$, Hang Xu$^{4}$, Haoqi Yuan$^{5,6}$, Xinrun Xu$^{2}$,}\\
  \textbf{Yu Zang$^{4}$, Liwei Hu$^{4}$, Junpeng Yue$^{5}$, Zhenxiong Jiang$^{4}$, Pengbo Hu$^{4}$,}\\
  \textbf{Börje F. Karlsson$^{3}$, Yehui Tang$^{4}$\textsuperscript{*}\thanks{\textsuperscript{*} Project Lead.}, Zongqing Lu$^{5,6}$}\textsuperscript{\dag}\thanks{\textsuperscript{\dag} Correspondence to Zongqing Lu <zongqing.lu@pku.edu.cn>.} \\
$^1$Institute of Automation, Chinese Academy of Sciences \\
$^2$University of Chinese Academy of Sciences \\
$^3$Beijing Academy of Artificial Intelligence, $^4$AgiBot \\
$^5$School of Computer Science, Peking University, $^6$BeingBeyond \\
}

\begin{document}

\maketitle
\thispagestyle{empty}
\pagestyle{empty}

%%%%%%%%%%%%%%%%%%%%%%%%%%%%%%%%%%%%%%%%%%%%%%%%%%%%%%%%%%%%%%%%%%%%%%%%%%%%%%%%
\begin{abstract}

In recent years, Multimodal Large Language Models (MLLMs) have demonstrated the ability to serve as high-level planners, enabling robots to follow complex human instructions. However, their effectiveness, especially in long-horizon tasks involving dual-arm humanoid robots, remains limited. This limitation arises from two main challenges: (i) the absence of simulation platforms that systematically support task evaluation and data collection for humanoid robots, and (ii) the insufficient embodiment awareness of current MLLMs, which hinders reasoning about dual-arm selection logic and body positions during planning. To address these issues, we present \projnamebold, a new dual-arm humanoid simulator, with continuous transition and a contingency mechanism. Building on this platform, we propose Proprio-MLLM, a model that enhances embodiment awareness by incorporating proprioceptive information with motion-based position embedding and a cross-spatial encoder. Experiments show that, while existing MLLMs struggle in this environment, Proprio-MLLM achieves an average improvement of 19.75\% in planning performance. Our work provides both an essential simulation platform and an effective model to advance embodied intelligence in humanoid robotics. The code is available at \url{https://anonymous.4open.science/r/DualTHOR-5F3B/}.

\end{abstract}

%%%%%%%%%%%%%%%%%%%%%%%%%%%%%%%%%%%%%%%%%%%%%%%%%%%%%%%%%%%%%%%%%%%%%%%%%%%%%%%%

\section{Introduction}

In recent years, leveraging Multimodal Large Language Models (MLLMs) as high-level planners to enable robots to execute complex human instructions has become a prominent research direction in Embodied AI \cite{song2023llm, zhao2024see}. To this end, the research community is actively pursuing a deep synergy between MLLMs and highly versatile dual-arm humanoid robots, aiming to unlock their vast potential in fine-grained \cite{humanplus}, long-horizon \cite{dagplan}, and human-like collaborative tasks \cite{mandi2024roco}. However, advancing this research frontier faces a fundamental bottleneck stemming from both data and models: first, a severe scarcity of high-level planning simulators for dual-arm humanoid robots that can support training data collection and evaluation of MLLMs; and second, the inherent capability limitations of existing MLLMs for dual-arm humanoid robot high-level planning tasks.

\begin{figure}[!t]
    \centering %表示居中
    \includegraphics[width=\columnwidth]{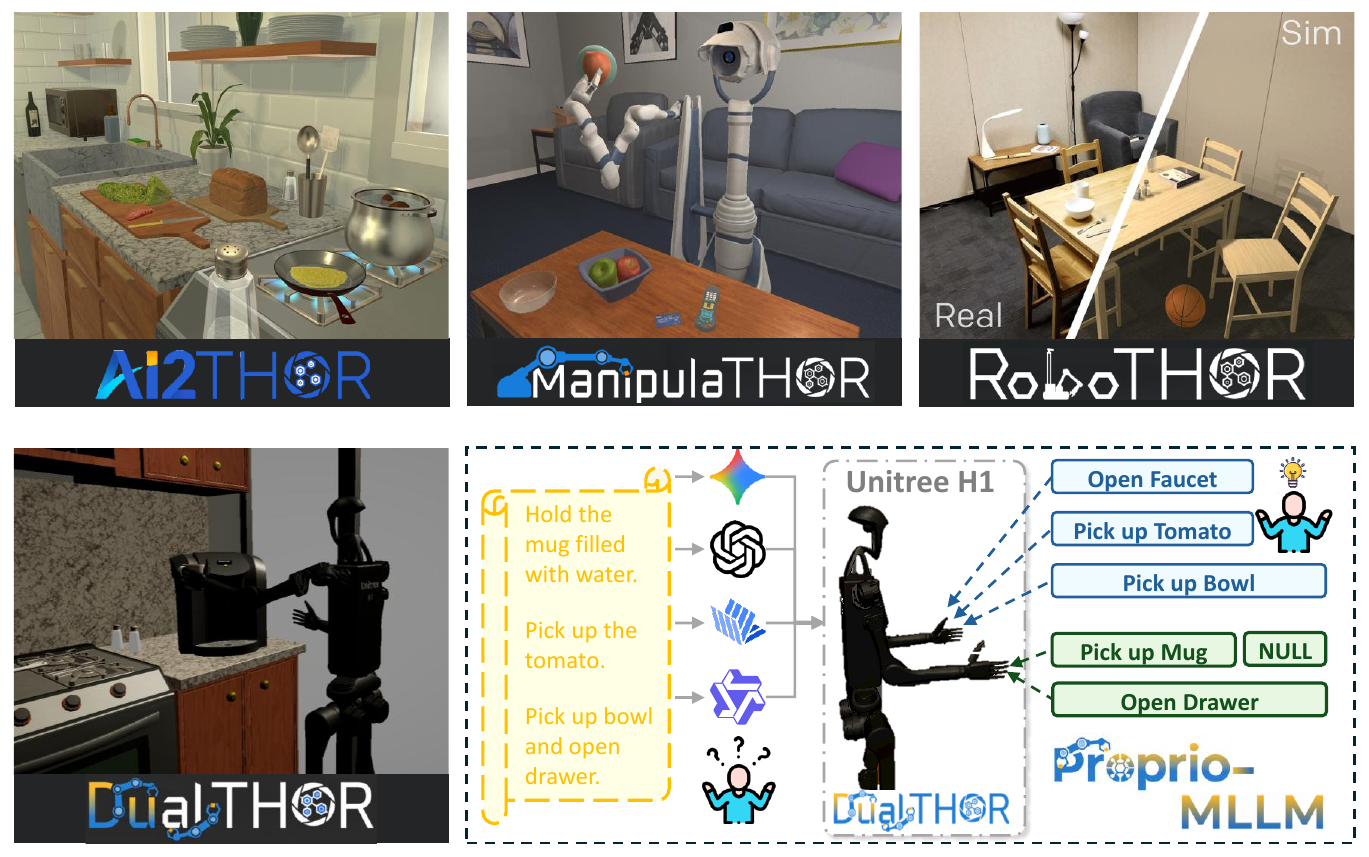}
    \vspace{-3mm}
    \caption{\projnamebold\ is a novel simulator specifically tailored for dual-arm humanoid robots, while still preserving the diversity and realism of scenarios in previous AI2-THOR series simulators. As current MLLMs have limited effectiveness in planning for dual-arm embodied tasks, we propose \textbf{Proprio-MLLM} to achieve proprioception-aware, embodiment-grounded planning.}
    \vspace{-8mm}
    \label{overview}
\end{figure}

This high-level planning data gap arises from systemic limitations in the current benchmarks \cite{robothor, procthor}, which fail to provide the long-horizon, visual-physically realistic humanoid interaction planning data required for MLLM training \cite{ding2025humanoid}. Specifically, this deficiency manifests in two aspects. First, existing benchmarks for long-horizon task planning \cite{ai2thor, manipulathor} are predominantly confined to wheeled robots or single-arm manipulators, as illustrated in the upper part of Fig. \ref{overview}. Moreover, they often sacrifice the physical realism of low-level control to focus on high-level logical evaluation. They rely on idealized, instantaneous state transitions \cite{zhang2025momakitchen100kbenchmarkaffordancegrounded} to ensure flawless skill sequencing, but this simplification produces data that is detached from physical reality, limiting the real-world applicability of MLLM planning policies and their sim-to-real transferability. Second, although some benchmarks support bimanual humanoid robots, they concentrate almost entirely on low-level control \cite{sferrazza2024humanoidbench, issacSim}, neglecting long-horizon planning challenges. Their evaluation scope is narrow, generating only fragmented data on isolated sub-skills such as picking and placing, without providing coherent task sequences necessary to assess performance in complex, multi-stage daily activities \cite{alfred}. Consequently, there is an urgent need for a benchmark platform capable of producing high-quality humanoid datasets that integrate finer-grained low-level control with long-horizon planning, filling this critical gap in the field.

Beyond the data scarcity in the current benchmarks, the second fundamental challenge lies within the MLLMs themselves. Although these models show encouraging capabilities in long-horizon embodied planning tasks \cite{kagaya2024rap}, their effectiveness is severely constrained by the lack of grounding in the dual-arm humanoid robots' physical reality \cite{luo2023perpetual}, as shown in the lower part of Fig. \ref{overview}. By failing to incorporate robots' embodiment, MLLMs often struggle with arm selection logic, body state adjustments (e.g., height changes), and the identification of coherent and physically plausible interaction points with objects \cite{lemma}. For instance, sometimes the target object is located on the left side of the robot, but the left hand is already occupied based on the MLLM's prior plan, while the right hand cannot reach it. Such conflicts typically necessitate re-planning for revised locomotion and manipulation, significantly reducing task success rates within a reasonable number of steps. Also, MLLMs should consider the robots' interaction range based on their joint configurations to find proper interaction points \cite{zhang2025momakitchen100kbenchmarkaffordancegrounded}, as low-level skills must always operate within a constrained range to guarantee stable execution. These limitations emphasize the importance of enhancing MLLMs' understanding of robot's embodiment.

To solve the aforementioned challenges, we first introduce a dual-arm humanoid simulation platform, \projnamebold, to support data collection and task evaluation for MLLMs. We employ dual-arm humanoid robots as primary agents and design a set of dual-arm tasks in which the two arms can either execute distinct actions in parallel or collaborate to accomplish a single complex task. We optimize the control logic to ensure continuous robot’s states and environmental transitions, and introduce a contingency mechanism to further simulate real-world uncertainty. Building on this foundation, We further propose \textbf{Proprio-MLLM}, an MLLM that incorporates proprioceptive information into dual-arm planning. We introduce a motion-based position embedding method and a cross-spatial encoder to enhance the model’s embodiment awareness and spatial reasoning abilities. Experimental results show that while existing MLLMs struggle with dual-arm planning tasks, Proprio-MLLM achieves an average improvement of 19.75\% in planning performance, demonstrating more reliable logical and spatial reasoning capabilities.

In summary, our contributions encompass the following key advancements:
\begin{enumerate}[left=1em, itemsep=1pt]
    % \vspace{-2mm}
    % \setlength{\itemsep}{1pt} 
    % \setlength{\leftskip}{0em}
    % \setlength{\itemindent}{0pt}
    \item We propose a dual-arm humanoid robot simulator \projnamebold\ (based on \aithor) and introduce a task suite for dual-arm planning. We create a new benchmark tailored for household dual-arm tasks, providing a standardized evaluation framework for future research.
    \item By incorporating proprioceptive information, we propose a multimodal alignment large language model, \textbf{Proprio-MLLM}. We introduce a motion-based position embedding method and a cross-spatial encoder, increasing the model’s embodiment awareness and spatial reasoning in dual-arm tasks.
    % We introduce humanoid IK functions and improved physical rendering techniques, realizing contingency and failures similar to real-world robotics to enhance the realism of the simulation environment.
    % Our simulator enable both high-level planning and faithful low-level execution, realizing contingency and failures similar to real-world robotics. Additionally, we introduce humanoid IK functions and improved physical rendering techniques to enhance the realism of the simulation environment.%, aimed at testing the generalization and robustness of MLLM planning.
    \item Our experimental results show that existing MLLMs have limited capabilities for dual-arm embodied planning tasks and Proprio-MLLM achieves an average improvement of 19.75\% in planning performance.
    %Further data expansion is needed to enhance embodied agents' understanding of dual-arm robot structures and task execution planning.
    %\vspace{-2mm}
\end{enumerate}

\section{Related Work}
\begin{table*}[!t]
    \centering
    \caption{A systematic comparison of \projname\ and existing long-horizon household simulation platforms.}
    \vspace{-1mm}
    \begin{adjustbox}{width=\textwidth, center}
    \setlength{\tabcolsep}{2mm}
    \renewcommand{\arraystretch}{0.9}
    \begin{tabular}{l l c c c c}
        \toprule
        \textbf{Simulator} & \textcolor{darkblue}{\textbf{Category}} & \textcolor{mydarkgreen}{\textbf{Agents}} & \textcolor{mybrown}{\textbf{Transition}} & \textcolor{mypink}{\textbf{Action Control}} & \textcolor{myblue2}{\textbf{Contingency}} \\
        \midrule
        ThreeDWorld \cite{gan2020threedworld} & Household & Wheel & Discrete & Discrete & \xmark \\
        iGibson \cite{li2021igibson} & Navigation & Wheel & Discrete & Discrete & \xmark \\
        AI2-THOR \cite{ai2thor} & Household & Wheel & Discrete & Discrete & \xmark \\
        RoboThor \cite{robothor} & Navigation & Wheel & Discrete & Discrete & \xmark \\
        ManipulaThor \cite{manipulathor} & Manipulation & Single Arm & Discrete & Discrete & \xmark \\
        ProcThor \cite{procthor} & Household & Wheel & Discrete & Discrete & \xmark \\
        OmniGibson \cite{li2023behavior} & Multi-Domain & Single Arm \& Wheel & Discrete & Discrete & \xmark \\
        MoMa-Kitchen \cite{zhang2025momakitchen100kbenchmarkaffordancegrounded} & Manipulation & Single Arm & Discrete & Discrete & \xmark \\
        \midrule
        \textbf{\projname\ (ours)} & Household & \cellcolor{tablepeach}{Dual Arm} & \cellcolor{tablepeach}{Continuous} & \cellcolor{tablepeach}{Discrete \& Continuous} & \cmark \\
        \bottomrule
    \end{tabular}
    \end{adjustbox}
    \label{comparison}
    \vspace{-6mm}
\end{table*}
\subsection{Simulation Platforms for Interactive Learning}

Simulation environments are indispensable for advancing interactive learning in embodied AI, offering controllable settings for agent training and evaluation. \aithor \cite{ai2thor} serves as a foundational framework, providing high-quality visual environments and interactive object dynamics that support a wide range of embodied AI tasks. Extensions such as RoboTHOR \cite{robothor}, ManipulaTHOR \cite{manipulathor}, and ProcTHOR \cite{procthor} have enhanced the simulator by increasing task diversity, scene complexity, and object variety, yet the range of robot embodiments remains limited to single-arm and wheeled robots. HumanoidBench \cite{sferrazza2024humanoidbench} and Isaac Gym \cite{issacSim} focus primarily on training low-level control policies for humanoid robots. Consequently, their scenes assets are limited for assessing long-horizon planning capabilities for MLLMs. To provide the necessary simulation capabilities along with a wide range of task categories and extended robot morphologies, we introduce our \projname\ simulator.

% However, existing simulation environments fall short in addressing the nuanced motion control requirements of bimanual humanoid robots and often oversimplify the process of motion interaction. These simplifications hinder the learning and testing of complex motor skills, particularly in tasks demanding precise bimanual coordination. To overcome these limitations, we propose a novel simulation environment specifically designed to provide enhanced motion control and advanced bimanual coordination capabilities. Our platform supports the execution of more intricate and realistic interactive tasks, thereby addressing the critical gaps in current frameworks and advancing the development of sophisticated motor skills in humanoid robotics.

%-------------------------------------------------------------------------
\subsection{Multimodal Alignment for Large Language Models}
Recent research on alignment techniques in MLLMs can be broadly categorized into deep fusion and shallow fusion methods. Deep fusion approaches, such as Flamingo \cite{alayrac2022flamingo} and NVLM \cite{dai2024nvlm}, modify the LLM’s architecture by incorporating cross-attention layers and additional feedforward components to improve modality alignment. In contrast, shallow fusion methods, including LLaVA \cite{liu2023visual} and MotionGPT \cite{jiang2023motiongpt}, map multimodal latent features into the LLM’s embedding space using either MLP-based mappings or cross-attention mechanisms. Shallow fusion methods offer greater computational efficiency, making them ideal for integrating proprioceptive information into MLLMs in a resource-efficient manner.

\section{DualTHOR}
\label{DualTHOR}

We introduce \projname, a novel simulation platform for dual-arm humanoid robots extended from \aithor. \projname\ significantly advances embodied interaction complexity by introducing a new suite of long-horizon tasks designed specifically for bimanual humanoid robots. Crucially, and in contrast to existing platforms, it supports continuous, physically-grounded interactions and incorporates a stochastic contingency mechanism to enable MLLMs to develop and refine their re-planning abilities in response to execution errors, as shown in Table \ref{comparison}. With these enhancements, \projname\ provides a highly interactive and flexible platform for advancing the development of robust dual-arm embodied systems.

% We introduce \projname, a simulator built upon \aithor \cite{ai2thor} and extended to dual-arm humanoid robots. The environment integrates the Unity physics engine \cite{haas2014history}, humanoid-specific inverse kinematics (IK) functions \cite{habekost2024inverse}, and task-reversal capabilities. With these capabilities, we design a novel set of tasks tailored specifically for dual-arm robots, significantly increasing the complexity of embodied interactions in household settings. A key feature of our platform is its support for continuous interaction processes, as shown in Table \ref{comparison}, distinguishing it from existing simulation platforms that often rely on discrete or simplified transitions. Moreover, to model real-world uncertainty, the environment incorporates a stochastic contingency mechanism, enabling agents to develop and refine their re-planning abilities in response to execution errors. With these enhancements, \projname\ provides a highly interactive and flexible platform for advancing the development of robust dual-arm embodied systems.

\subsection{Overview of DualTHOR}

\begin{figure}[h]
  \centering
  \begin{subfigure}[b]{0.45\columnwidth}
    \centering
    \includegraphics[width=\textwidth]{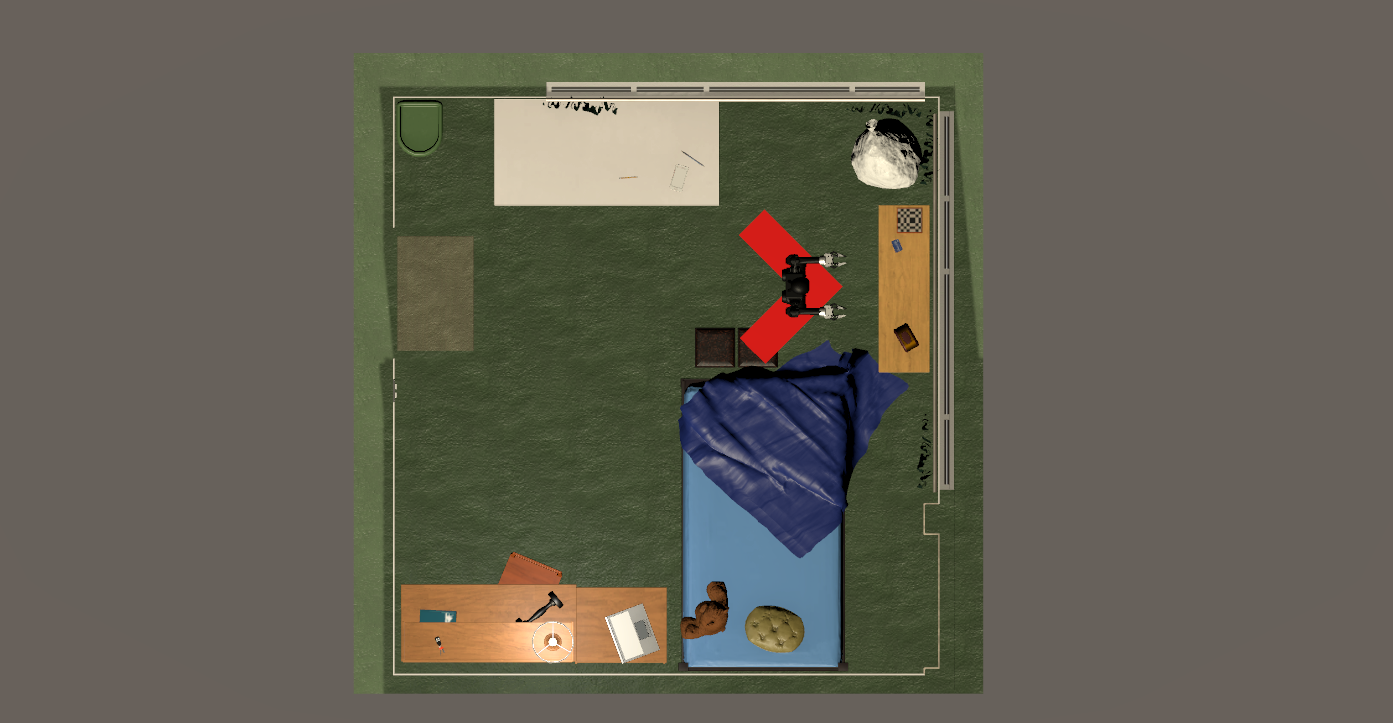}
    \vspace{-0.5cm}
    \caption{Bedroom}
  \end{subfigure}
  \begin{subfigure}[b]{0.45\columnwidth}
    \centering
    \includegraphics[width=\textwidth]{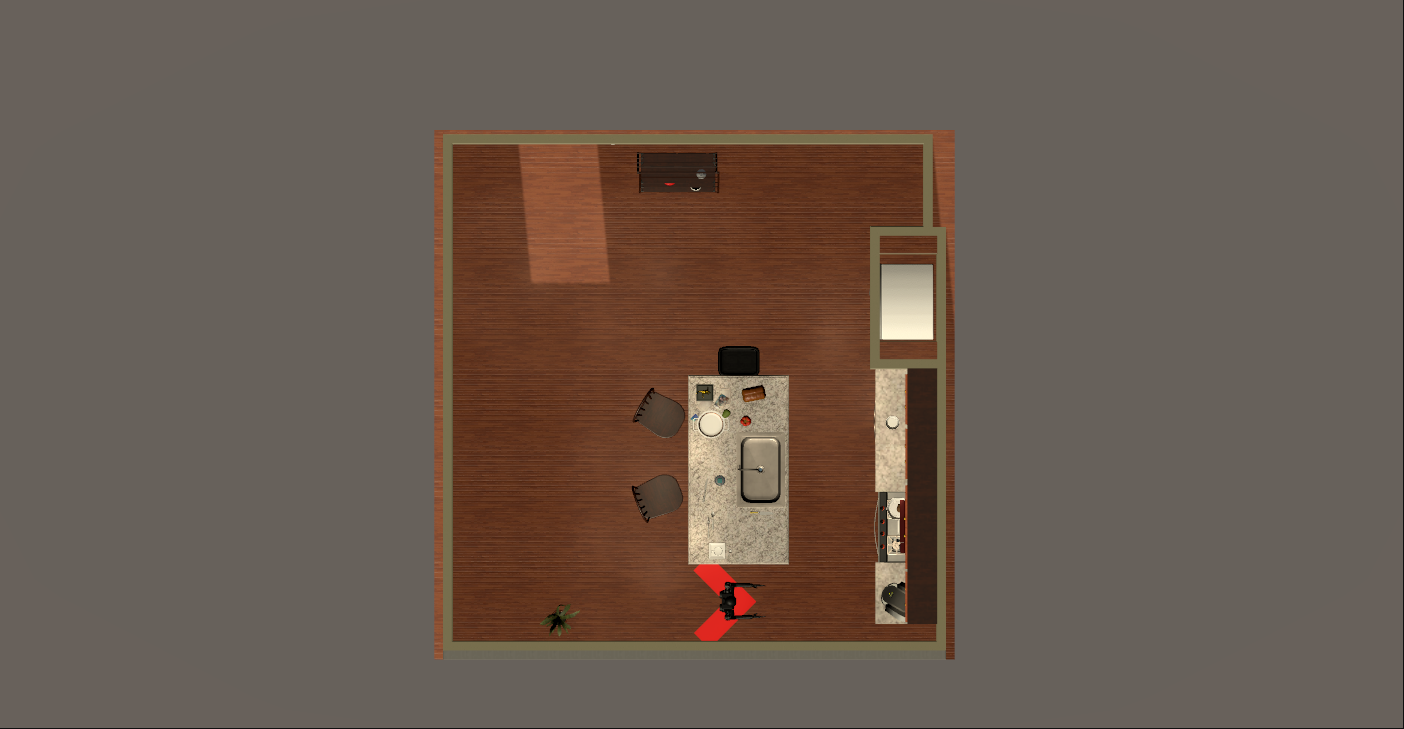}
    \vspace{-0.5cm}
    \caption{Kitchen}
  \end{subfigure}
  \begin{subfigure}[b]{0.45\columnwidth}
    \centering
    \includegraphics[width=\textwidth]{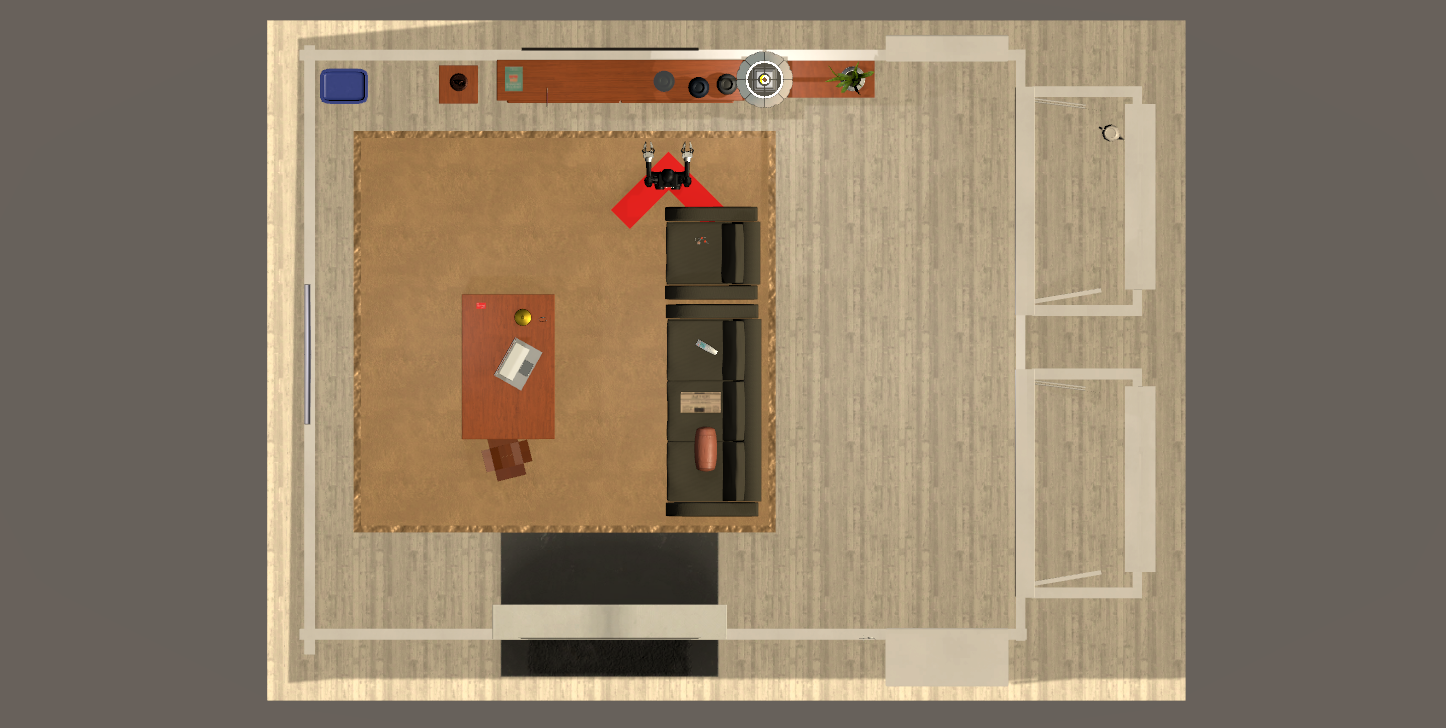}
    \vspace{-0.5cm}
    \caption{Living room 1}
  \end{subfigure}
  \begin{subfigure}[b]{0.45\columnwidth}
    \centering
    \includegraphics[width=\textwidth]{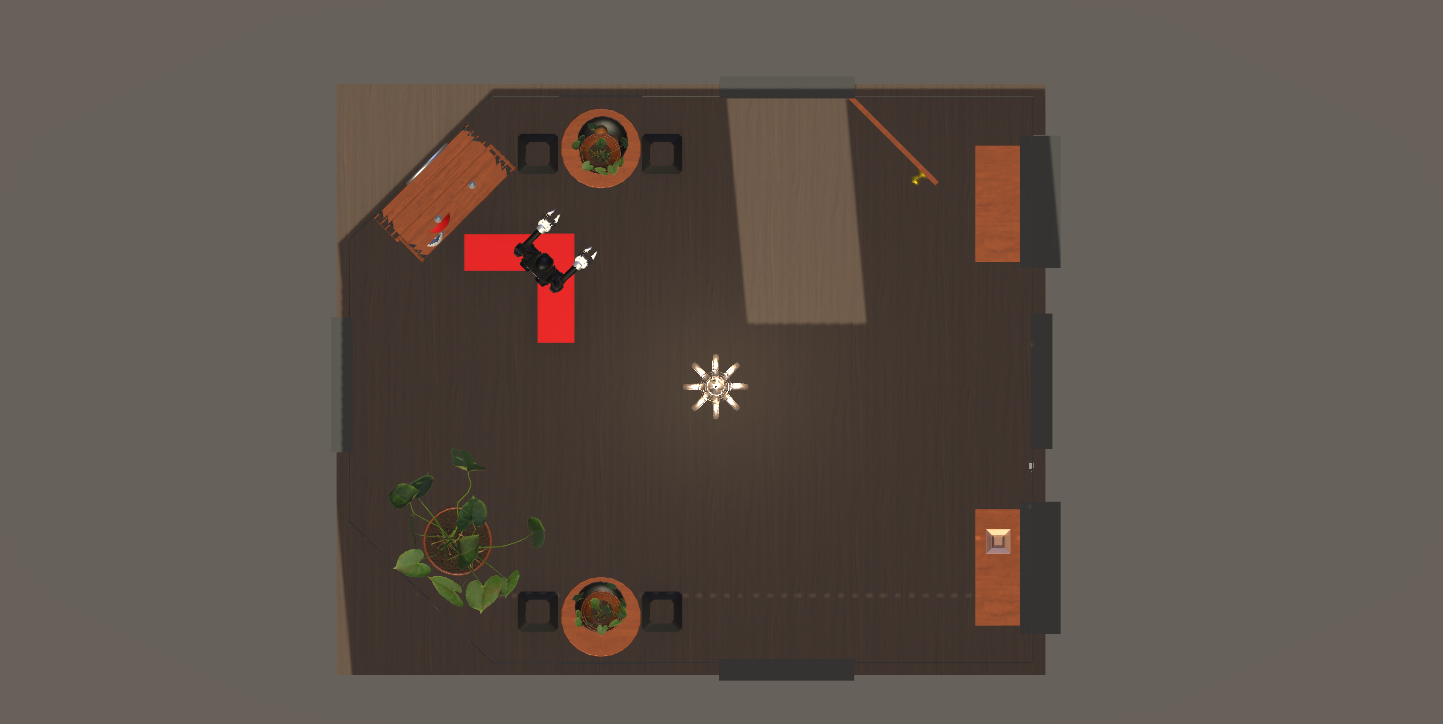}
    \vspace{-0.5cm}
    \caption{Living room 2}
  \end{subfigure}
  \vspace{-0.1cm}
  \caption{\textbf{Example scenes of different rooms in \projname.} The types and quantities of objects vary across rooms, and the humanoid robot is capable of interacting with all objects within each room.}
  \label{scene}
  \vspace{-4mm}
\end{figure}

\textbf{Physics Engine.} We build \projname\ on the Unity engine to support physically realistic bimanual manipulation. Unity's framework facilitates the parallel execution of coordinated arm actions and ensures smooth, continuous transitions through interpolation. This physical and visual fidelity supported by our platform is critical for perception-driven agents operating in the diverse and complex household scenarios. \projname\ is a multi-camera suite designed to provide comprehensive situational awareness across diverse household scenes (see Fig. \ref{scene}) while mitigating self-occlusion.

% \textbf{Visual Perception.} 
% A key feature of \projname is a multi-camera suite designed to provide comprehensive situational awareness across diverse household scenes (see Figure~\ref{scene}) while mitigating self-occlusion. In addition to standard first-person cameras on the robot's head, the platform includes a suite of third-person cameras that dynamically adjust their position to follow the robot's movements, minimizing occlusion from its own limbs. This system provides both egocentric and exocentric views of the agent's interaction with the environment, which is crucial for complex manipulation and navigation tasks.

\textbf{Humanoid Robots.} To support a broad range of interaction scenarios, \projname\ integrates two distinct humanoid robots: the powerful and stable Unitree H1, and the dexterous and precise Agibot X1. This diversity enables research on tasks ranging from heavy-duty interactions (H1) to fine-motor control (X1). Crucially, the two robots feature different end-effectors—dexterous hands for H1 and parallel grippers for X1—allowing the study of different manipulation paradigms within a unified platform.

\subsection{Task Categories}
\label{taskcategory}
Tasks in \projname\ are designed to rigorously evaluate bimanual capabilities. To this end, we introduce a novel categorization based on the required number of arms:
\begin{itemize}[left=1em]
\item \textbf{Dual-Arm Essential Tasks:} Tasks that are physically impossible with a single arm. Examples include lifting heavy objects or coordinating to hold an object while opening an affordance-specific container.
\item \textbf{Dual-Arm Optional Tasks:} Tasks that can be performed with a single arm, but are enhanced through dual-arm execution, such as transporting two objects within constrained timesteps.
\item \textbf{Single-Arm Tasks:} Simple interaction tasks, largely adapted from \aithor, that serve as basic plan tasks.
\end{itemize}

\subsection{Low-Level Control}

Low-level control in \projname\ operates on a client-server model, where a Python API sends high-level commands to the Unity engine. For interaction actions, Unity queries a modular inverse kinematics (IK) service that uses OmniManip \cite{omnimanip} to compute the required joint configurations. This solver is particularly suited for our platform due to its support for dexterous hand models, and the resulting trajectories are smoothed to ensure continuous, realistic motion.

A key aspect of our low-level control is the implementation of distinct IK models tailored to each robot's specific architecture. The Agibot X1 adopts a decoupled approach, solving the IK for each arm independently using a simplified pose matrix (a 3×3 rotation matrix with a translation vector) relative to its base frame. In contrast, the Unitree H1 utilizes a whole-body coordination model that solves for both arms simultaneously. This model employs a complete 4×4 homogeneous transformation matrix to incorporate full posture information and integrates current joint states (angles and velocities) for dynamic optimization, with a specific focus on maintaining balance constraints during bimanual operations. This dual-model approach allows \projname\ to faithfully simulate the distinct control paradigms of different real-world humanoid systems.

\subsection{Contingency Mechanism}
\begin{figure}[h]
    \centering %表示居中
    \includegraphics[width=0.96\linewidth]{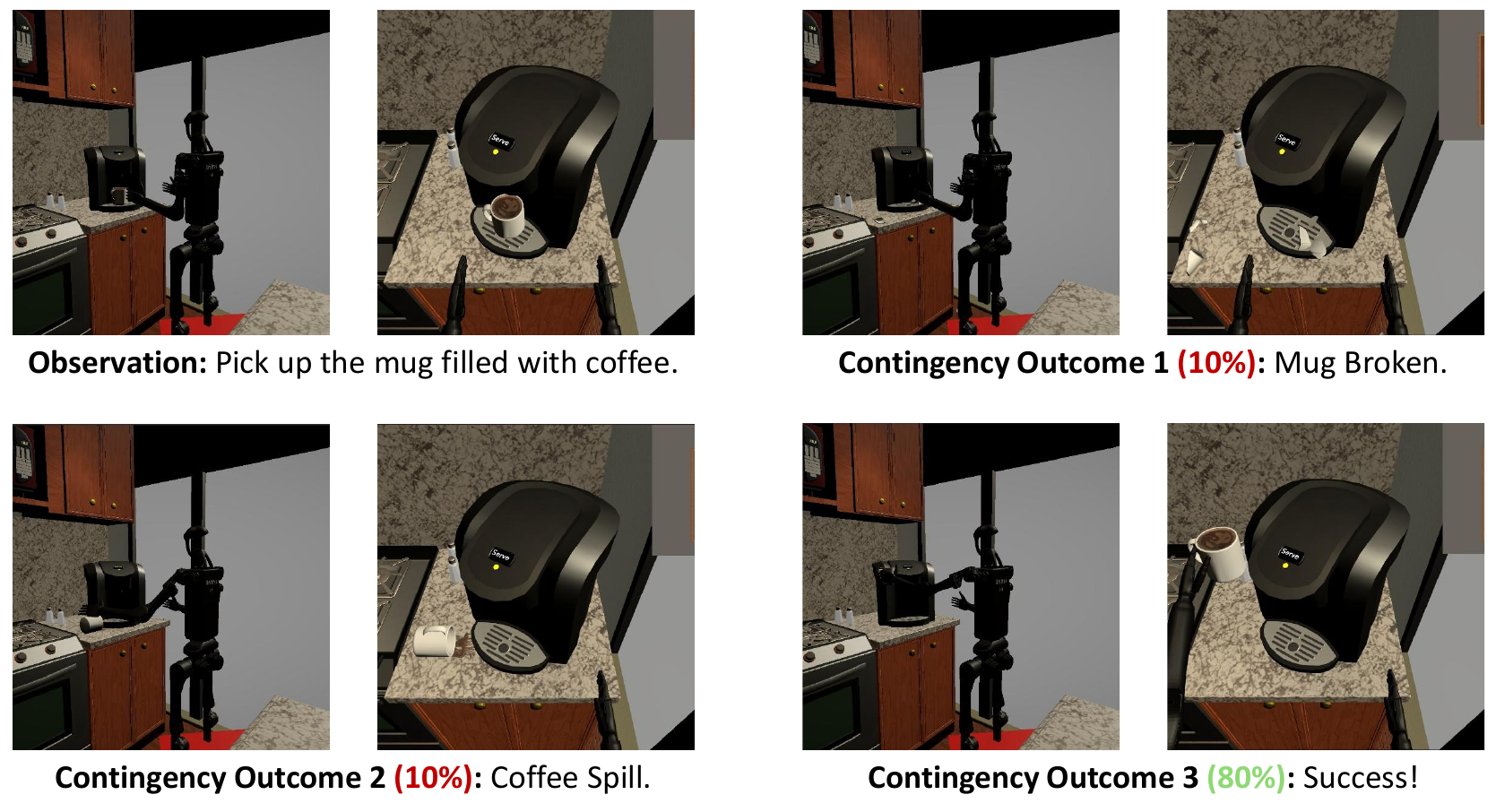}
    \caption{\textbf{Example of picking up a ``\textit{pourable}" cup of coffee.} The possible results include success (80\%), coffee spill (10\%), and mug broken (10\%). \projname\ provides both visual observations and environmental feedback after the robot executes an action, enabling the evaluation of the effectiveness of the current plan and the acquisition of information necessary for MLLM re-planning.}
    \label{pickup_mug}
    \vspace{-6mm}
\end{figure}

We introduce a stochastic contingency mechanism in \projname. This system is designed to simulate real-world uncertainty, by mapping an action to a distribution of potential outcomes rather than a single guaranteed result. For example, when an agent attempts to pick up a "\textit{pourable}" cup, there is an 80\% probability of success, but also a 10\% chance of the cup breaking and a 10\% chance of its contents spilling (illustrated in Fig. \ref{pickup_mug}). These stochastic outcomes, derived from object categories, allow the agent to move beyond rigid plans and require MLLMs to develop robust error-recovery and re-planning capabilities. 

% Furthermore, to enhance realism and logical consistency, actions in our system are constrained by the current state of objects. For instance, an ingredient labeled as “Cooked” cannot undergo another “COOK” action, as it has reached a terminal state for that process. However, it may remain “\textit{pickupable},” allowing for continued manipulation. This state-dependent execution model prevents unrealistic action sequences and supports the multi-state object representations necessary to capture the complexity of real-world, long-horizon tasks.

% Existing frameworks, such as \aithor, typically assume deterministic state transitions following action execution, which fails to capture the inherent complexity and uncertainty of real-world interactions. This deterministic assumption introduces a substantial distributional gap between simulated environments and the real-world, thereby constraining the robustness and adaptability of trained agents. In scenarios where actions fail or yield unintended consequences, agents often lack the capacity for dynamic re-planning or error recovery, ultimately resulting in task failure. To address this limitation, we propose a probabilistic contingency mechanism designed to simulate the stochastic nature of physical environments, thereby enhancing agents’ robustness and adaptability under uncertainty.

\section{Method}
\begin{figure*}[!t]
    \centering %表示居中
    \includegraphics[width=\textwidth]{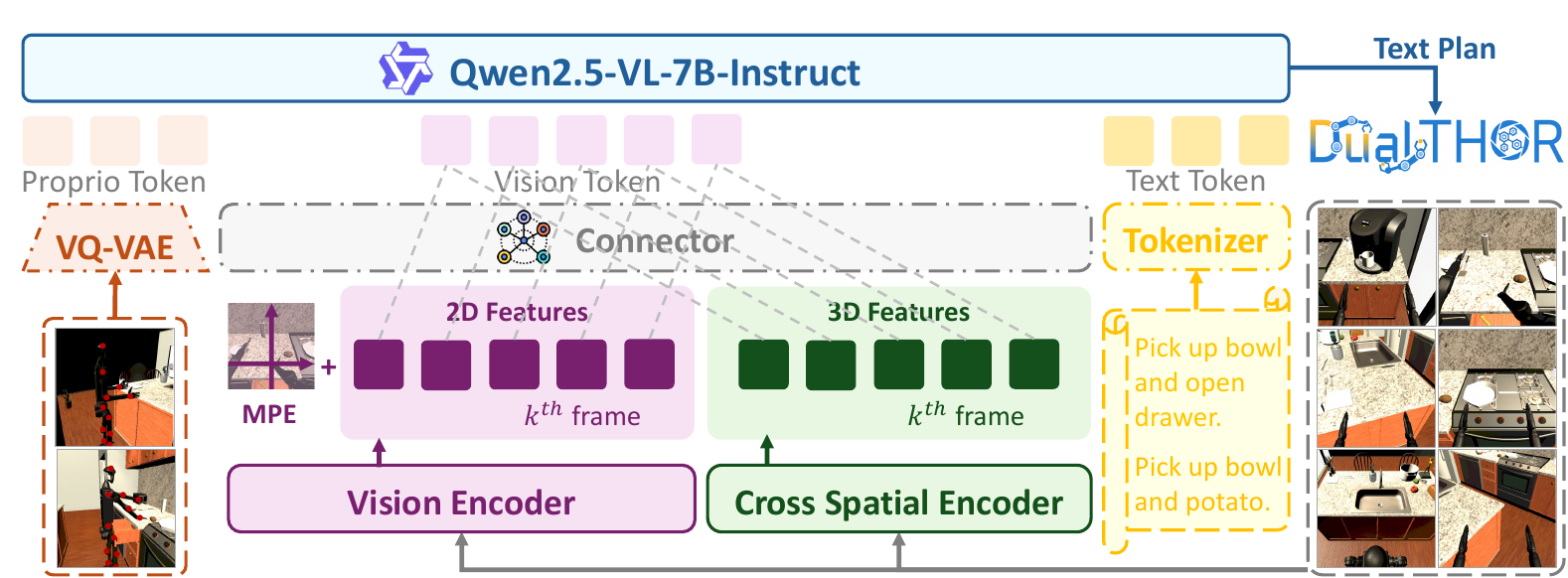}
    \vspace{-2mm}
    \caption{\textbf{The framework of Proprio-MLLM.} By incorporating proprioceptive information, we propose a multimodal alignment large language model, \textbf{Proprio-MLLM}. We introduce a motion-based position embedding method and a cross-spatial encoder, increasing the model’s embodiment awareness and spatial reasoning in dual-arm tasks.}
    \vspace{-6mm}
    \label{method}
\end{figure*}

In this work, we propose \textbf{Proprio-MLLM}, a model for long-horizon dual-arm robot planning that tackles two key challenges: (i) integrating proprioceptive information into the planning process to resolve arm selection logic and body state changes, and (ii) enhancing spatial reasoning to ensure coherent and physically plausible interaction points with objects. To achieve this, we extend Qwen 2.5-VL-7B-Ins \cite{bai2025qwen2} with multi-modal motion inputs, enabling it to perceive and reason over the robot’s proprioceptive states. We introduce a motion-based position embedding and a cross-spatial encoder for proprioceptive information grounding. The overall architecture is shown in Fig. \ref{method}.

% In this work, we propose \textbf{Proprio-MLLM}, a framework designed to address two fundamental challenges in long-horizon planning for dual-arm robots: (i) integrating proprioceptive information into the planning process to resolve arm selection logic and manage body state changes (e.g., height adjustments), and (ii) enhancing spatial reasoning to ensure coherent and physically plausible interaction points with objects. To this end, we extend a MLLM by incorporating multi-modal motion inputs, enabling it to perceive and reason over the robot’s proprioceptive states. Specifically, we introduce a motion-based image position embedding that captures local spatio-temporal cues relevant to arm selection and body configuration changes. Furthermore, to provide the model with a better understanding of the robot’s interaction range, we design an image spatial cross encoder that improves spatial reasoning and interaction grounding. The overall model architecture is illustrated in Figure 3.

\subsection{Alignment Dataset Preparation}
Motivated by MotionGPT \cite{jiang2023motiongpt}, we use motion modal to help MLLMs understand the proprioceptive states of dual-arm humanoid robots. We build a robot motion-image-text dataset upon the HumanML3D dataset \cite{guo2022generating}, which provides different human action samples for motion generation tasks. As the HumanML3D dataset provides humanoid motion in SMPL format \cite{loper2023smpl}, we retarget the human motion data into the H1, X1 robots and replay them in DualTHOR simulator to obtain corresponding first-person visual observations. For consistency with the output of spatial encoder, robot motions are reformulated as joint positions. After PHC-based filtering \cite{luo2023perpetual}, we obtain 11k high-quality trajectories for training.
% For consistency with the spatial encoder output, we reformulate the robot’s motion data as joint positions. We further applied the PHC algorithm \cite{luo2023perpetual} for data filtering, resulting in approximately 11k high-quality trajectories for alignment training.

\subsection{Motion Alignment Training}
We separate the training scheme into three stages. First, we learn the motion tokenizer to align with the MLLM's token-in-token-out format. Second, we fine-tune the MLLM to achieve motion, language and image feature alignment. Finally, we conduct instruction tuning to enable Proprio-MLLM to solve embodied dual-arm tasks.

\textbf{Stage 1: Motion Tokenizer. }We introduce a motion VQ-VAE \cite{van2017neural} model, which consists of a motion encoder $E$, a motion decoder $D$, and a codebook $C = \{c_1, c_2, \dots, c_M\}$ of size $M$. We denote a robot motion sequence from alignment dataset as $m = \{m_1, m_2, \dots, m_T\} \in \mathbb{R}^{T \times d}$, where $T$ is the sequence length and $d$ is the feature dimension per frame. The encoder $E$ projects $m$ into a latent embedding $z(m)$, which is then discretized by quantization. Specifically, each embedding is mapped to its nearest entry in the learnable codebook $C$. The quantized vector $e$ and its code index $p$ are defined as:
\begin{equation}
e = c_p, \quad
p = \arg \min_k \| z(m) - c_k \|_2 .
\end{equation}

The motion VQ-VAE is trained with three tailored loss: a reconstruction loss, a codebook loss and a commitment loss.
The overall loss is formulated as:
\begin{equation}
\begin{split}
\mathcal{L} = & \ \|D(z(m)) - m\|^2 \\
& + \alpha \| sg[Z(m)] - e \|_2^2 
+ \beta \| Z(m) - sg[e] \|_2^2 ,
\end{split}
\end{equation}
where $\text{sg}[\cdot]$ denotes the stop-gradient operation and $\alpha, \beta$ are hyper-parameters to control the relative weight of different components. The codebook is updated using an Exponential Moving Average (EMA) strategy to improve stability. 

\textbf{Stage 2: Alignment Training.} We combine all modalities into a unified token space, utilizing shared embedding layers and attention mechanisms. The attention mechanisms of concatenated hidden states $H_{m,i,t} = [H_m, H_i, H_t]$ including motion, image and text embeddings are calculated by, 
% \begin{equation}
%     Q_{m,i,t} = W_Q H_{m,i,t}, K_{m,i,t} = W_K H_{m,i,t}, V_{m,i,t} = W_V H_{m,i,t},
% \end{equation}
\begin{equation}
\begin{aligned}
Q_{m,i,t} &= W_Q H_{m,i,t}, \quad K_{m,i,t} = W_K H_{m,i,t}, \\
V_{m,i,t} &= W_V H_{m,i,t}.
\end{aligned}
\end{equation}
where $\{Q, K, V\}_{m,i,t}$ represents query, key and value, $W_{\{Q, K, V\}}$ represents their weight matrices.

In this stage, we fine-tune the MLLM with Low-Rank Adaptation (LoRA) \cite{hu2022lora} to capture complex interdependencies across different modalities, such as understanding the robot body state when interacting with the objects or adjusting the body height when text requires. Since MLLMs already exhibit strong image-text alignment, we initially freeze the image and text weights separately to improve motion alignment of single modality to prevent treating motion as noise. After training stabilizes, we perform joint training across all modalities.

\textbf{Stage 3: Instruction Tuning.}
Finally, we use a self-learning method SELU \cite{li2024selu} to collect some dual-arm planning data, therefore grounding the MLLM into dual-arm planning tasks with the understanding of all three modalities.

\subsection{Motion-Based Position Embedding}
To better integrate embodiment information and strengthen the correlation between motion and visual inputs, thereby enhancing Proprio-MLLM’s long-horizon reasoning based on the robot’s body state, we introduce a motion-based position embedding (MPE) for visual tokens. Prior methods such as M-RoPE \cite{wang2024qwen2} encode temporal, height, and width positions, where temporal IDs remain constant for image and spatial IDs are assigned by pixel location. However, this design is less effective for embodied tasks, as it ignores interactive regions around the robot and their relative spatial relationships.

To overcome this limitation, we replace the original dimensions (temporal, height, width) with (temporal, robot-centric height, robot-centric width), where spatial indices are encoded radially outward from the robot’s centroid. Specifically, for a visual token at $(x, y)$ and a robot centroid at $(x_r, y_r)$, we define:
\begin{equation}
MPE(t, x, y) = \big[t,\;\; \text{sign}(y - y_r),\;\; \text{sign}(x - x_r)\big],
\end{equation}
where $t$ denotes the temporal index. The MPE is added to visual tokens, enabling the model to distinguish tokens directions relative to the robot.

\begin{figure*}[h]
  \centering
  \begin{subfigure}[b]{0.27\textwidth}
    \centering
    \includegraphics[width=\textwidth]{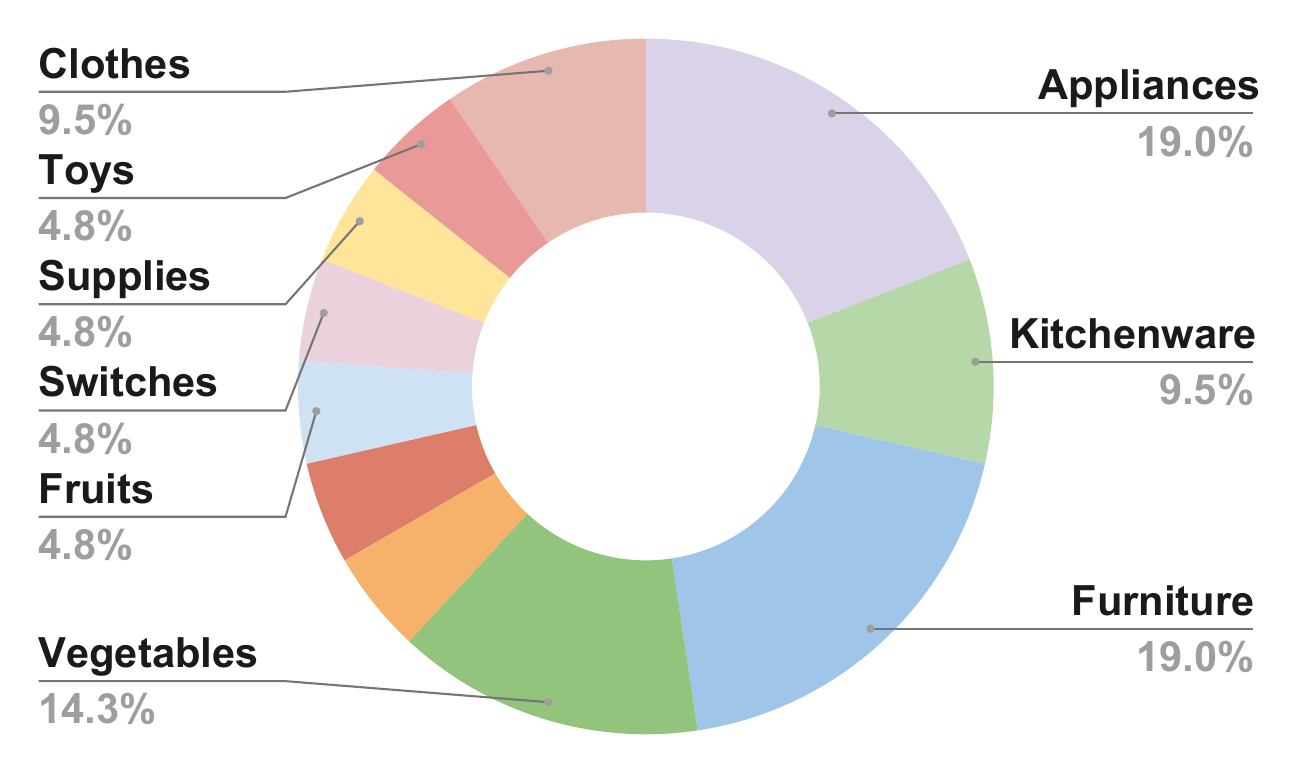}
    \vspace{-0.1cm}
    \caption{Distribution of interactive object types in \projname.}
    \label{object}
  \end{subfigure}
  \hfill
  \begin{subfigure}[b]{0.27\textwidth}
    \centering
    \includegraphics[width=\textwidth]{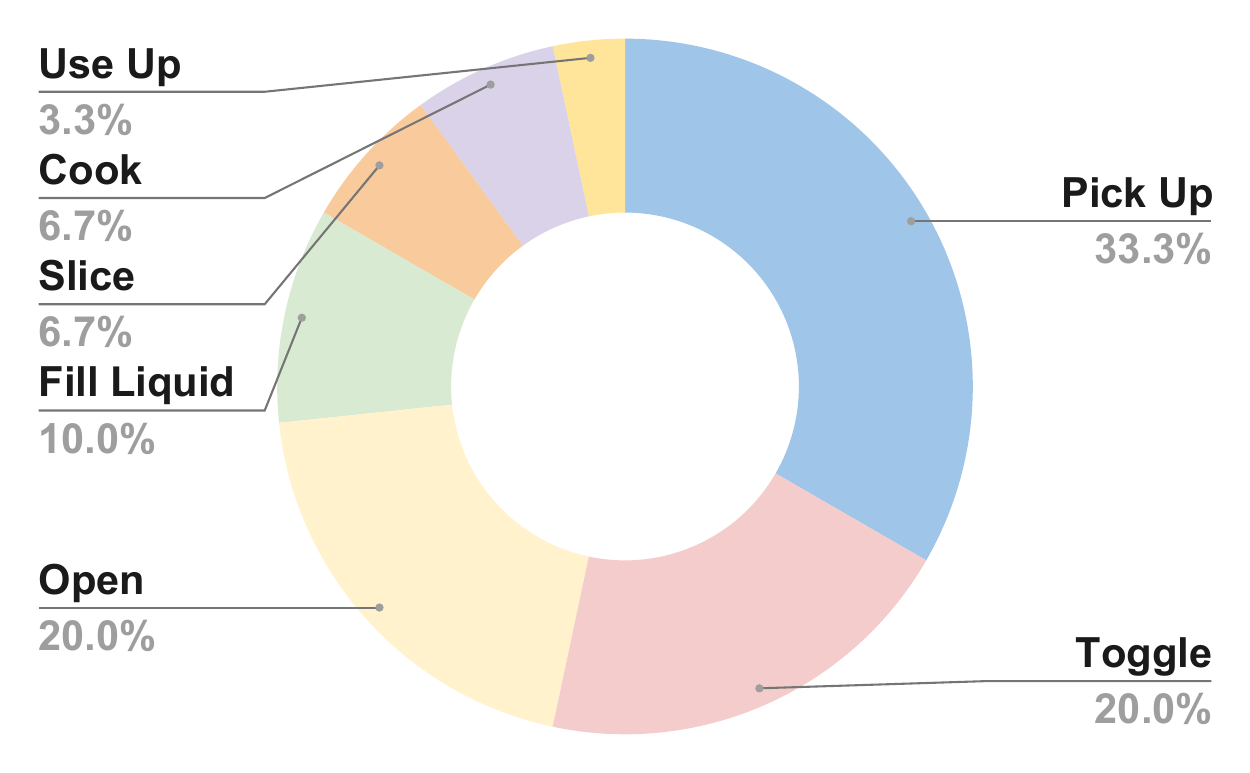}
    \vspace{-0.1cm}
    \caption{Distribution of single-arm interaction tasks in \projname.}
    \label{task}
  \end{subfigure}
  \hfill
  \begin{subfigure}[b]{0.27\textwidth}
    \centering
    \includegraphics[width=\textwidth]{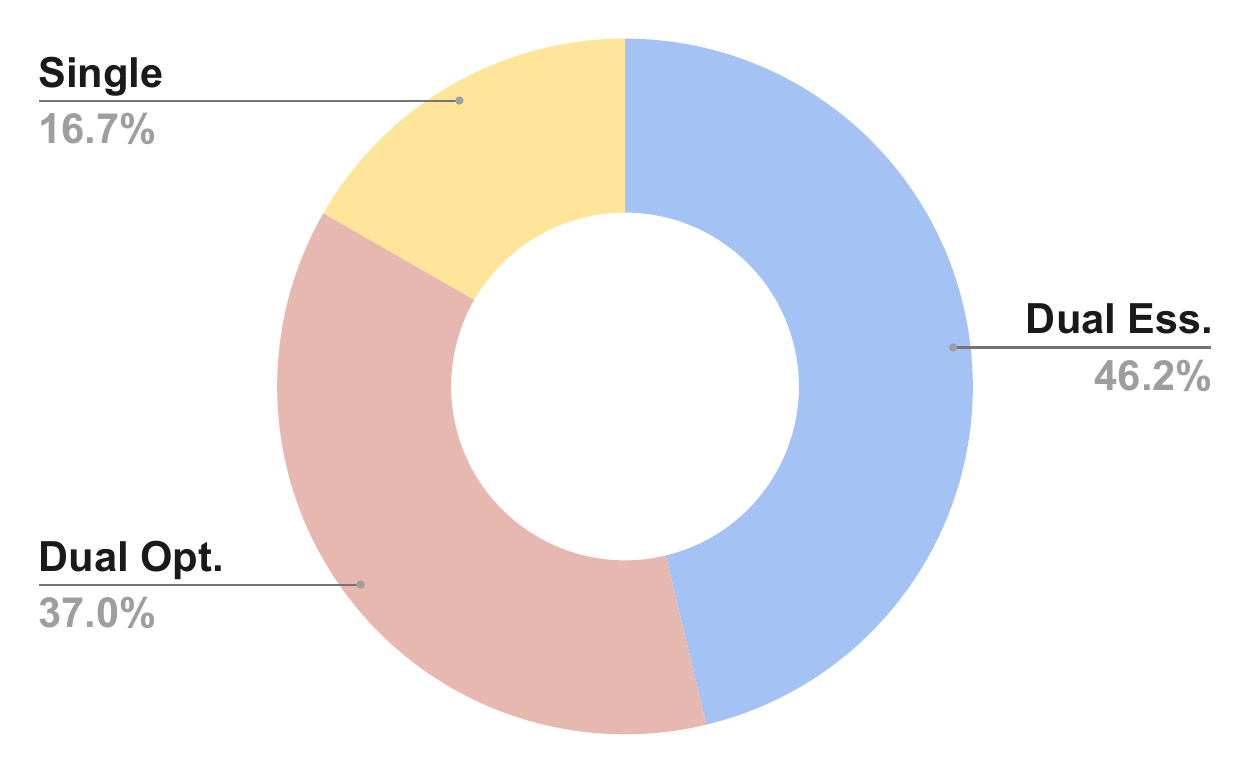}
    \vspace{-0.1cm}
    \caption{Distribution of the three task categories in \projname.}
    \label{category}
  \end{subfigure}
  \caption{\textbf{Distribution of objects and tasks.}}
  \label{demo}
  \vspace{-4mm}
\end{figure*}

\begin{table*}[h]
\caption{\textbf{Performance of baselines across different task configurations in \projname.} Each task includes results for both X1 and H1 robots. The evaluation metric is the success rate over 50 trials for each task.}
\vspace{-1mm}
\centering
\begin{adjustbox}{width=0.9\textwidth, center}
\begin{tabular}{ccccccc} 
\toprule
 & \multicolumn{2}{c}{\textbf{Dual-Arm Essential Tasks}} & \multicolumn{2}{c}{\textbf{Dual-Arm Optional Tasks}} & \multicolumn{2}{c}{\textbf{Single-Arm Tasks}} \\
\cmidrule(lr){2-3} \cmidrule(lr){4-5} \cmidrule(lr){6-7}
 & X1 & H1 & X1 & H1 & X1 & H1 \\
\midrule
\textbf{GPT-4o} & 23.31\% & 27.07\% & 39.76\% & 40.96\% & 51.67\% & 56.67\% \\
\textbf{Gemini-1.5-Pro} & 25.56\% & 26.33\% & 37.35\% & 36.14\% & 41.67\% & 43.33\% \\
\textbf{Qwen2.5-VL-7B-Ins} & 18.05\% & 17.29\% & 21.08\% & 19.28\% & 23.33\% & 25.00\% \\
\textbf{InternVL2.5-8B} & 9.77\% & 15.79\% & 23.49\% & 25.30\% & 23.33\% & 31.67\% \\
\textbf{LLM-Planner} & 28.57\% & 31.58\% & 43.37\% & 45.78\% & 55.00\% & 56.67\% \\
\textbf{RAP} & 27.81\% & 33.51\% & 45.18\% & 47.59\% & 53.33\% & 55.00\% \\
\textbf{DAG-Plan} & 36.09\% & 41.53\% & 51.20\% & 52.41\% & 55.00\% & 58.33\% \\
\midrule
\textbf{Proprio-MLLM} & \textbf{59.39\%} & \textbf{63.16\%} & \textbf{71.69\%} & \textbf{70.48\%} & \textbf{73.33\%} & \textbf{75.00\%} \\
\bottomrule 
\end{tabular}
\end{adjustbox}
%\vspace{2mm}
\label{performance}
\vspace{-6mm}
\end{table*}

\subsection{Cross-Spatial Encoder}

While proprioceptive integration and motion-based position embedding allow Proprio-MLLM to better capture dual-arm selection logic and adapt planning to the current embodiment, a limitation remains in estimating the robot’s interactive range. Although the model can identify task-relevant objects and adjust posture, it still struggles to determine whether an object is within feasible reach.

To address this, we introduce a cross-spatial encoder (CSE) to enhance interactive range prediction. Compared with VGGT \cite{wang2025vggt}, CUT3R \cite{wang2025continuous} provides more accurate single-frame depth estimation, yielding reliable geometric cues when combined with arm-length constraints. Specifically, we retain the Qwen2.5-VL image encoder for high-level visual features and augment it with 3D point-map features predicted by CUT3R from a single RGB image. Given an observation $o_i \in \mathbb{R}^{H \times W \times 3}$, we obtain:
\begin{equation}
F_{\text{2D}} = E_{\text{Qwen2.5-VL}}(o_i), \quad
F_{\text{3D}} = E_{\text{CUT3R}}(o_i),
\end{equation}
where $F_{\text{2D}}$ denotes 2D visual features and $F_{\text{3D}}$ represents the point-map based features. The 3D features are reshaped, aligned with 2D features, and fused via a lightweight MLP:
\begin{equation}
T_{\text{vision}} = \text{MLP}\big[(F_{2D} + \Phi(F_{3D}))],
\end{equation}
where $\Phi(\cdot)$ reshapes 3D features into the 2D-compatible space. The final fused vision token $T_{\text{vision}}$, combined with proprioceptive states, provides enhanced awareness of the robot’s interactive range, enabling more accurate dual-arm interaction planning.

\section{Experiments}
\subsection{Experimental Setup}
\label{section 4.1}
We conduct experiments on 359 tasks across 10 distinct room environments, involving 68 unique objects. Distributions of object categories, task categories, and the proportion of each task type are illustrated in Fig. \ref{demo}. Most dual-arm essential tasks and dual-arm optional tasks are composed of multiple single-arm tasks, while a smaller portion is manually designed. 
% More details of all tasks are described in Appendix~\ref{Appendix B}. 

\subsection{Baselines}
We evaluate three distinct classes of baselines in \projname: proprietary MLLMs, open-source MLLMs, and prompt-enhanced MLLMs. The planning capabilities of these models for dual-arm tasks are assessed mainly by their success rates across all task categories.

Proprietary MLLMs (GPT-4o \cite{hurst2024gpt}, Gemini 1.5 Pro \cite{team2023gemini}) are closed-source commercial systems trained on large-scale private datasets and optimized for downstream tasks. Open-source MLLMs (Qwen2.5-VL-7B \cite{bai2025qwen2}, InternVL2.5-8B \cite{chen2024expanding}) are publicly available models, typically trained on transparent datasets. Prompt-enhanced MLLMs (LLM Planner \cite{song2023llm}, RAP \cite{kagaya2024rap}, DAG-Plan \cite{dagplan}) leverage structured prompts to improve reasoning in embodied planning. Base model used is GPT-4o. DAG-Plan \cite{dagplan} is the state-of-the-art method for dual-arm planning that utilizes a graph-based structure to represent dual-arm skills.

\subsection{Main Results for Dual-Arm Tasks}
\label{section 4.3}

We test all baselines across the three task categories defined in Section \ref{taskcategory}. Each task is tested on both X1 and H1 robots with 50 trials to ensure statistical reliability. Initial positions vary across rooms but remain consistent within the same room to preserve the reliability of historical trajectories for prompt-enhanced MLLMs. All baselines are tested under identical conditions: temperature set to 0, maximum token length 2048, input images scaled to $500 \times 500$ resolution, and a maximum of 50 environment steps for high-level planning.

\textbf{Is proprioceptive information necessary for long-horizon dual-arm planning tasks?} Our results in Table~\ref{performance} show that existing methods struggle with long-horizon dual-arm planning. By incorporating proprioceptive information along with the proposed motion-based position embedding and cross-spatial encoder, Proprio-MLLM achieves the best overall performance. Compared to DAG-Plan, which leverages graph structures to represent robot skills, Proprio-MLLM improves average performance by 19.75\%, and up to 23.30\% on dual-arm essential tasks for X1 robot. Across the three task categories, we observe a pronounced gap between dual-arm and single-arm planning in current MLLMs, highlighting the scarcity of dual-arm humanoid planning data in existing simulators and motivating the development of \projname\ for more diverse embodied planning data. Notably, proprioceptive information consistently enhances performance across all tasks, with the largest gains observed in dual-arm scenarios.

\begin{figure}[h]
    \centering %表示居中
    \includegraphics[width=\columnwidth]{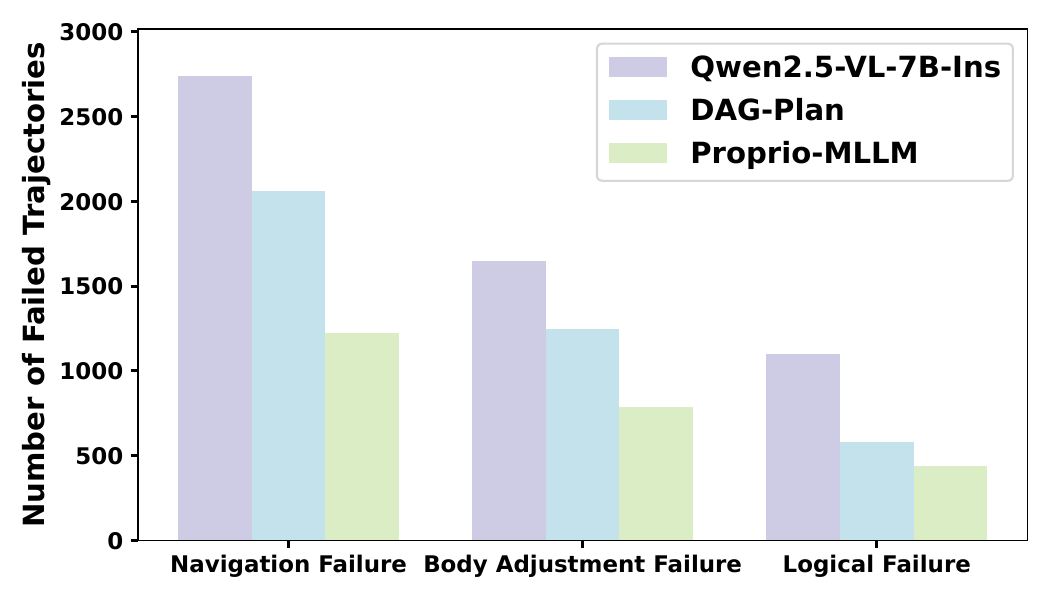}
    \vspace{-6mm}
    \caption{\textbf{Impact of proprioceptive integration.} We analyze the failed trajectories of the basic model, the state-of-the-art method, and our Proprio-MLLM, categorizing these failures into three types to assess how each model leverages proprioceptive information.}
    \label{failure}
    \vspace{-4mm}
\end{figure}

\textbf{How does proprioceptive information influence dual-arm high-level planning?} To highlight the impact of proprioceptive information on high-level planning, we analyze failure trajectories and categorize them into three types: navigation failure (unable to reach the target interaction point), body adjustment failure (mismatched height with the target object leading to low-level execution failure), and logical failure (suboptimal arm selection in long-horizon planning). Statistics in Fig. \ref{failure} show that both DAG-Plan and Proprio-MLLM reduce failures, indicating that proprioceptive information improves arm selection logic. However, DAG-Plan’s graph-based prompting cannot fully exploit proprioception for other failure types. This limitation motivates the use of motion-based position embedding and the cross-spatial encoder, which together enable Proprio-MLLM to integrate image perception and proprioceptive states more effectively, producing coherent and adaptive dual-arm plans.

\vspace{-2mm}
\begin{table}[h]
\caption{\textbf{Performance of re-planning in \projname.} Tasks are categorized into three difficulty levels based on the success rate of low-level skills.}
\vspace{-1mm}
\centering
\begin{adjustbox}{width=0.9\columnwidth, center}
\begin{tabular}{c ccc}
\toprule
\multirow{2}{*}{\textbf{Method}} 
& \multicolumn{3}{c}{\textbf{H1}} \\
\cmidrule(lr){2-4}
 & Easy & Medium & Hard \\
\midrule
Proprio-MLLM    & \textbf{63.16\%} & \textbf{51.63\%} & \textbf{36.51\%} \\
Proprio (w/o reflection)    & 56.39\% & 38.67\% & 26.17\% \\
DAG-Plan    & 41.53\% & 20.33\% & 15.53\% \\
DAG (w/o reflection)    & 34.81\% & 10.97\% & 7.26\% \\
\bottomrule
\end{tabular}
\end{adjustbox}
\label{robustness}
\vspace{-3mm}
\end{table}

\textbf{How effectively can MLLMs handle contingency?} Since \projname\ provides both continuous interaction and a contingency mechanism, our benchmark can evaluate the re-planning ability of MLLMs, which is fundamental to ensuring the safety of high-level planning in real-world deployment. Each task is further divided into three difficulty levels — Easy (100\% action success rate), Medium (50\%), and Hard (20\%). To accomplish such tasks, agents must analyze failure scenarios and re-plan accordingly. For example, they should locate a new cup when the original one is broken. We evaluate the performance of DAG-Plan and Proprio-MLLM, with and without the reflection prompt, across three categories, and report the representative results on H1 dual-arm essential tasks in Table~\ref{robustness}. From the experimental results, we observe that Proprio-MLLM leverages proprioceptive information more effectively, and the reflection prompt contributes more substantially to improving re-planning ability. However, the limited performance of both methods on high-difficulty tasks reveals the scarcity of contingency cases in the training data of current MLLMs. This limitation further underscores the importance of \projname\ in providing a contingency mechanism for the development of robust planning datasets.

\vspace{-2mm}
\begin{table}[h]
\caption{\textbf{Results of ablation study of different components in Proprio-MLLM.} We examine the impact of MPE, CSE, and proprioceptive integration by analyzing the reduction of three failure categories.}
% \vspace{-1mm}
\centering
\begin{adjustbox}{width=0.9\columnwidth, center}
\renewcommand{\arraystretch}{0.95}
\begin{tabular}{c c ccc}
\toprule
\multirow{2}{*}{\textbf{Method}} & \multirow{2}{*}{\textbf{Success Rate}} & \multicolumn{3}{c}{\textbf{Failure Categories}} \\
\cmidrule(lr){3-5}
& & \#1 & \#2 & \#3 \\
\midrule
Proprio-MLLM & \textbf{63.16\%} & \cellcolor{tablepeach}{1225} & \cellcolor{tableblue}{784} & \cellcolor{tableblue}{441} \\
Proprio w/o CSE & 45.17\% & \cellcolor{tablepeach}{2205} & 902 & 538 \\
Proprio w/o MPE & 52.33\% & 1585 & \cellcolor{tableblue}{951} & \cellcolor{tableblue}{634} \\
Proprio w/o ALL & 41.53\% & 2238 & 972 & 678 \\
Qwen2.5-VL-Ins & 17.29\% & 2751 & 1650 & 1100 \\
\bottomrule
\end{tabular}
\end{adjustbox}
\label{ablation}
\vspace{-5mm}
\end{table}

\subsection{Ablation Study}

To investigate the impact of each component in \textbf{Proprio-MLLM}, we conduct ablation experiments by removing the motion-based position embedding (Proprio w/o MPE) and cross-spatial encoder (Proprio w/o CSE), as shown in Table \ref{ablation}. Proprio w/o ALL refers to removing both components, isolating the effect of proprioceptive information integration. Failure \#1–3 correspond to navigation, body adjustment, and logical failures. The results show that incorporating proprioceptive information significantly enhances MLLM planning by aligning with the robot's body state. Specifically, Proprio w/o CSE emphasizes the cross-spatial encoder's role in improving the robot’s interaction range, reducing navigation failures (orange squares). In contrast, Proprio w/o MPE highlights the importance of motion-based position embedding in adjusting the robot’s body, minimizing arm selection logical errors and reducing the latter two failure types (blue squares). Combining both components enables Proprio-MLLM to achieve proprioception-aware, long-horizon planning for dual-arm humanoid robots.

% \subsection{Realistic Environment Transitions}
% \label{section4.4}
% In addition to the contingency mechanism designed to mimic real-world environmental non-determinism, \projname\ further enhances realism by continuously updating the rendering of scenes. As illustrated in Figure~\ref{engine}, the simulator includes dynamic changes such as the gradual filling of a sink after turning on the faucet. This feature provides a more immersive and realistic environment, enabling MLLMs to better comprehend scene transformations and assess whether planned actions are being or have been successfully executed. By incorporating such fine-grained physical interactions, \projname\ not only improves the fidelity of the simulation, but also challenges MLLMs to adapt their action plans in response to continuous environmental changes, thereby advancing their applicability in real-world scenarios.
% \begin{figure*}[h]
%     \centering %表示居中
%     \includegraphics[width=0.8\linewidth]{img/fill_water.png}
%     \caption{Rendered observations showing a sink gradually filling up with water. \projname\ uses improved physics rendering techniques to provide agents with detailed action effects. % increase MLLM's judgment of the execution of actions.
%     }
%     \label{engine}
%     \vspace{-5mm}
% \end{figure*}

\section{Conclusion and Future Work}
We present \projname, a high-fidelity planning simulation environment designed to enable continuous and realistic bimanual humanoid robot interaction. We design three categories of dual-arm tasks to evaluate planning ability of MLLMs. To optimize the dual-arm planning performance, we propose Proprio-MLLM to utilize the proprioceptive information into long-horizon plan. We introduce motion-based position embedding and cross-spatial encoder to understand the dual-arm humanoid body state and interaction range to achieve proprioception-aware plan and reduce three kinds of failures. Proprio-MLLM can achieve an average improvement of 19.75\% in planning performance, demonstrating more reliable logical and spatial reasoning capabilities.

Flexible asset generation tools and support for a broader range of robots are under development. Further refinements to support more controllable failure modes, and multi-room environments are also planned. Additionally, future work should incorporate multi-agent cooperative evaluation.% and integrate reinforcement learning tools for low-level policy learning, especially for humanoid robot control.

% In future work, we plan to further expand our simulation environment by incorporating a broader range of 3D assets and robots. To support scalability and diversity, future work can develop autonomous asset generation tools capable of producing a wide variety of objects and scenes. Additionally, future work should incorporate multi-agent evaluation and integrate reinforcement learning tools for low-level policy learning, especially for humanoid robot control. %to facilitate the optimization of humanoid control policies.

\bibliographystyle{IEEEtran}
\bibliography{IEEEabrv}

\begin{thebibliography}{10}
\providecommand{\url}[1]{#1}
\csname url@rmstyle\endcsname
\providecommand{\newblock}{\relax}
\providecommand{\bibinfo}[2]{#2}
\providecommand\BIBentrySTDinterwordspacing{\spaceskip=0pt\relax}
\providecommand\BIBentryALTinterwordstretchfactor{4}
\providecommand\BIBentryALTinterwordspacing{\spaceskip=\fontdimen2\font plus
\BIBentryALTinterwordstretchfactor\fontdimen3\font minus \fontdimen4\font\relax}
\providecommand\BIBforeignlanguage[2]{{%
\expandafter\ifx\csname l@#1\endcsname\relax
\typeout{** WARNING: IEEEtran.bst: No hyphenation pattern has been}%
\typeout{** loaded for the language `#1'. Using the pattern for}%
\typeout{** the default language instead.}%
\else
\language=\csname l@#1\endcsname
\fi
#2}}

\bibitem{song2023llm}
C.~H. Song, J.~Wu, C.~Washington, \emph{et~al.}, ``Llm-planner: Few-shot grounded planning for embodied agents with large language models,'' in \emph{Proceedings of the IEEE/CVF international conference on computer vision}, 2023, pp. 2998--3009.

\bibitem{zhao2024see}
Z.~Zhao, W.~Chai, X.~Wang, \emph{et~al.}, ``See and think: Embodied agent in virtual environment,'' in \emph{European Conference on Computer Vision}.\hskip 1em plus 0.5em minus 0.4em\relax Springer, 2024, pp. 187--204.

\bibitem{humanplus}
Z.~Fu, Q.~Zhao, Q.~Wu, \emph{et~al.}, ``Humanplus: Humanoid shadowing and imitation from humans,'' \emph{arXiv preprint arXiv:2406.10454}, 2024.

\bibitem{dagplan}
Z.~Gao, Y.~Mu, J.~Qu, \emph{et~al.}, ``Dag-plan: Generating directed acyclic dependency graphs for dual-arm cooperative planning,'' \emph{arXiv preprint arXiv:2406.09953}, 2024.

\bibitem{mandi2024roco}
Z.~Mandi, S.~Jain, and S.~Song, ``Roco: Dialectic multi-robot collaboration with large language models,'' in \emph{2024 IEEE International Conference on Robotics and Automation (ICRA)}, 2024.

\bibitem{robothor}
M.~Deitke, W.~Han, A.~Herrasti, \emph{et~al.}, ``Robothor: An open simulation-to-real embodied ai platform,'' \emph{2020 IEEE/CVF Conference on Computer Vision and Pattern Recognition}, 2020.

\bibitem{procthor}
M.~Deitke, E.~VanderBilt, A.~Herrasti, \emph{et~al.}, ``Procthor: Large-scale embodied ai using procedural generation,'' \emph{Advances in Neural Information Processing Systems}, 2022.

\bibitem{ding2025humanoid}
P.~Ding, J.~Ma, X.~Tong, \emph{et~al.}, ``Humanoid-vla: Towards universal humanoid control with visual integration,'' \emph{arXiv preprint arXiv:2502.14795}, 2025.

\bibitem{ai2thor}
E.~Kolve, R.~Mottaghi, W.~Han, \emph{et~al.}, ``Ai2-thor: An interactive 3d environment for visual ai,'' \emph{arXiv preprint arXiv:1712.05474}, 2017.

\bibitem{manipulathor}
K.~Ehsani, W.~Han, A.~Herrasti, \emph{et~al.}, ``Manipulathor: A framework for visual object manipulation,'' \emph{2021 IEEE/CVF Conference on Computer Vision and Pattern Recognition}, 2021.

\bibitem{zhang2025momakitchen100kbenchmarkaffordancegrounded}
P.~Zhang, X.~Gao, Y.~Wu, \emph{et~al.}, ``Moma-kitchen: A 100k+ benchmark for affordance-grounded last-mile navigation in mobile manipulation,'' \emph{arXiv preprint arXiv:2503.11081}, 2025.

\bibitem{sferrazza2024humanoidbench}
C.~Sferrazza, D.-M. Huang, X.~Lin, \emph{et~al.}, ``Humanoidbench: Simulated humanoid benchmark for whole-body locomotion and manipulation,'' \emph{arXiv preprint arXiv:2403.10506}, 2024.

\bibitem{issacSim}
V.~Makoviychuk, L.~Wawrzyniak, Y.~Guo, \emph{et~al.}, ``Isaac gym: High performance gpu-based physics simulation for robot learning,'' \emph{arXiv preprint arXiv:2108.10470}, 2021.

\bibitem{alfred}
M.~Shridhar, J.~Thomason, D.~Gordon, \emph{et~al.}, ``Alfred: A benchmark for interpreting grounded instructions for everyday tasks,'' \emph{Computer Vision and Pattern Recognition}, 2020.

\bibitem{kagaya2024rap}
T.~Kagaya, T.~J. Yuan, Y.~Lou, \emph{et~al.}, ``Rap: Retrieval-augmented planning with contextual memory for multimodal llm agents,'' in \emph{NeurIPS Workshop on Open-World Agents}, 2024.

\bibitem{luo2023perpetual}
Z.~Luo, J.~Cao, K.~Kitani, \emph{et~al.}, ``Perpetual humanoid control for real-time simulated avatars,'' in \emph{Proceedings of the IEEE/CVF International Conference on Computer Vision}, 2023.

\bibitem{lemma}
R.~Gong, X.~Gao, Q.~Gao, \emph{et~al.}, ``Lemma: Learning language-conditioned multi-robot manipulation,'' \emph{IEEE Robotics and Automation Letters}, 2023.

\bibitem{gan2020threedworld}
C.~Gan, J.~Schwartz, S.~Alter, \emph{et~al.}, ``Threedworld: A platform for interactive multi-modal physical simulation,'' \emph{arXiv preprint arXiv:2007.04954}, 2020.

\bibitem{li2021igibson}
C.~Li, F.~Xia, R.~Mart{\'\i}n-Mart{\'\i}n, \emph{et~al.}, ``igibson 2.0: Object-centric simulation for robot learning of everyday household tasks,'' \emph{arXiv preprint arXiv:2108.03272}, 2021.

\bibitem{li2023behavior}
C.~Li, R.~Zhang, J.~Wong, \emph{et~al.}, ``Behavior-1k: A benchmark for embodied ai with 1,000 everyday activities and realistic simulation,'' in \emph{Conference on Robot Learning}, 2023, pp. 80--93.

\bibitem{alayrac2022flamingo}
J.-B. Alayrac, J.~Donahue, P.~Luc, \emph{et~al.}, ``Flamingo: a visual language model for few-shot learning,'' \emph{Advances in neural information processing systems}, 2022.

\bibitem{dai2024nvlm}
W.~Dai, N.~Lee, B.~Wang, \emph{et~al.}, ``Nvlm: Open frontier-class multimodal llms,'' \emph{arXiv preprint arXiv:2409.11402}, 2024.

\bibitem{liu2023visual}
H.~Liu, C.~Li, Q.~Wu, \emph{et~al.}, ``Visual instruction tuning,'' \emph{Advances in neural information processing systems}, 2023.

\bibitem{jiang2023motiongpt}
B.~Jiang, X.~Chen, W.~Liu, \emph{et~al.}, ``Motiongpt: Human motion as a foreign language,'' \emph{Advances in Neural Information Processing Systems}, vol.~36, pp. 20\,067--20\,079, 2023.

\bibitem{omnimanip}
M.~Pan, J.~Zhang, T.~Wu, \emph{et~al.}, ``Omnimanip: Towards general robotic manipulation via object-centric interaction primitives as spatial constraints,'' in \emph{Proceedings of the Computer Vision and Pattern Recognition Conference}, 2025.

\bibitem{bai2025qwen2}
S.~Bai, K.~Chen, X.~Liu, \emph{et~al.}, ``Qwen2.5-vl technical report,'' \emph{arXiv preprint arXiv:2502.13923}, 2025.

\bibitem{guo2022generating}
C.~Guo, S.~Zou, X.~Zuo, \emph{et~al.}, ``Generating diverse and natural 3d human motions from text,'' in \emph{Proceedings of the IEEE/CVF conference on computer vision and pattern recognition}, 2022, pp. 5152--5161.

\bibitem{loper2023smpl}
M.~Loper, N.~Mahmood, J.~Romero, \emph{et~al.}, ``Smpl: A skinned multi-person linear model,'' in \emph{Seminal Graphics Papers: Pushing the Boundaries, Volume 2}, 2023, pp. 851--866.

\bibitem{van2017neural}
A.~Van Den~Oord, O.~Vinyals, \emph{et~al.}, ``Neural discrete representation learning,'' \emph{Advances in neural information processing systems}, vol.~30, 2017.

\bibitem{hu2022lora}
E.~J. Hu, Y.~Shen, P.~Wallis, \emph{et~al.}, ``Lora: Low-rank adaptation of large language models.'' \emph{ICLR}, vol.~1, no.~2, p.~3, 2022.

\bibitem{li2024selu}
B.~Li, H.~Jiang, Z.~Ding, \emph{et~al.}, ``Selu: Self-learning embodied mllms in unknown environments,'' \emph{arXiv preprint arXiv:2410.03303}, 2024.

\bibitem{wang2024qwen2}
P.~Wang, S.~Bai, S.~Tan, \emph{et~al.}, ``Qwen2-vl: Enhancing vision-language model's perception of the world at any resolution,'' \emph{arXiv preprint arXiv:2409.12191}, 2024.

\bibitem{wang2025vggt}
J.~Wang, M.~Chen, N.~Karaev, \emph{et~al.}, ``Vggt: Visual geometry grounded transformer,'' in \emph{Proceedings of the Computer Vision and Pattern Recognition Conference}, 2025, pp. 5294--5306.

\bibitem{wang2025continuous}
Q.~Wang, Y.~Zhang, A.~Holynski, \emph{et~al.}, ``Continuous 3d perception model with persistent state,'' in \emph{Proceedings of the Computer Vision and Pattern Recognition Conference}, 2025, pp. 10\,510--10\,522.

\bibitem{hurst2024gpt}
A.~Hurst, A.~Lerer, A.~P. Goucher, \emph{et~al.}, ``Gpt-4o system card,'' \emph{arXiv preprint arXiv:2410.21276}, 2024.

\bibitem{team2023gemini}
G.~Team, R.~Anil, S.~Borgeaud, \emph{et~al.}, ``Gemini: a family of highly capable multimodal models,'' \emph{arXiv preprint arXiv:2312.11805}, 2023.

\bibitem{chen2024expanding}
Z.~Chen, W.~Wang, Y.~Cao, \emph{et~al.}, ``Expanding performance boundaries of open-source multimodal models with model, data, and test-time scaling,'' \emph{arXiv preprint arXiv:2412.05271}, 2024.

\end{thebibliography}

\addtolength{\textheight}{-12cm}   % This command serves to balance the column lengths
                                  % on the last page of the document manually. It shortens
                                  % the textheight of the last page by a suitable amount.
                                  % This command does not take effect until the next page
                                  % so it should come on the page before the last. Make
                                  % sure that you do not shorten the textheight too much.

%%%%%%%%%%%%%%%%%%%%%%%%%%%%%%%%%%%%%%%%%%%%%%%%%%%%%%%%%%%%%%%%%%%%%%%%%%%%%%%%

%%%%%%%%%%%%%%%%%%%%%%%%%%%%%%%%%%%%%%%%%%%%%%%%%%%%%%%%%%%%%%%%%%%%%%%%%%%%%%%%

%%%%%%%%%%%%%%%%%%%%%%%%%%%%%%%%%%%%%%%%%%%%%%%%%%%%%%%%%%%%%%%%%%%%%%%%%%%%%%%

\end{document}